\documentclass{salesforce_ai_research}

\usepackage{amsmath}
\usepackage{amssymb}
\usepackage{graphicx}
\usepackage{booktabs}
\usepackage{lipsum}
\usepackage{tcolorbox}
\usepackage{cleveref} % Required for \cref
\usepackage{fontawesome5}
\usepackage{spverbatim}

\newcommand{\eg}[1]{\textit{e.g.}}
\newcommand{\ie}[1]{\textit{i.e.}}
\definecolor{sfblue}{HTML}{009EDB}
\newcommand{\blue}[1]{\textcolor{sfblue}{\textbf{#1}}}

\title{W\&D: Scaling Parallel Tool Calling for Efficient Deep Research Agents}

\author{
Xiaoqiang Lin\thanks{Equal Contributions.}\footnotemark[1] \quad Jun Hao Liew\footnotemark[1]\quad Silvio Savarese \quad Junnan Li \\
Salesforce AI Research \\[0.5em]
\faGithub\,\href{https://github.com/SalesforceAIResearch/MCP-Universe/tree/main/mcpuniverse/benchmark/configs/deepresearch}{SalesforceAIResearch/MCP-Universe/deepresearch} \\[0.2em]
\faGlobe\, \href{https://xqlin98.github.io/wide-deep-research-agent/}{Website}
}

\begin{document}

\maketitle
\begin{abstract}
    Deep research agents have emerged as powerful tools for automating complex intellectual tasks through multi-step reasoning and web-based information seeking. While recent efforts have successfully enhanced these agents by scaling depth through increasing the number of sequential thinking and tool calls, the potential of scaling width via parallel tool calling remains largely unexplored. In this work, we propose the \textbf{Wide and Deep research agent}, a framework designed to investigate the behavior and performance of agents when scaling not only depth but also width via parallel tool calling. Unlike existing approaches that rely on complex multi-agent orchestration to parallelize workloads, our method leverages intrinsic parallel tool calling to facilitate effective coordination within a single reasoning step. We demonstrate that scaling width significantly improves performance on deep research benchmarks while reducing the number of turns required to obtain correct answers. 
    Furthermore, we analyze the factors driving these improvements through case studies and explore various tool call schedulers to optimize parallel tool calling strategy. Our findings suggest that optimizing the trade-off between width and depth is a critical pathway toward high-efficiency deep research agents. Notably, without context management or other tricks, we obtain 62.2\% accuracy with GPT-5-Medium on BrowseComp, surpassing the original 54.9\% reported by GPT-5-High.  
\end{abstract}

\section{Introduction}

Deep research agents (\citep{openai2025deepresearch,gemini2025deepresearch,team2025tongyi,liu2025deepseek,kimi2025researcher,team2025mirothinker}) have raised increasing interest in both applications and the research community due to their growing capabilities and potential to conduct multi-step research on the internet, liberating humans from complex intellectual tasks. These agents perform multi-step reasoning and information seeking to solve tasks that typically require hours of human work. Specifically, at each step, these agents perform reasoning followed by tool execution to search and retrieve information from the web. Ultimately, they provide a final answer or a full report that summarizes all the information gathered after multiple steps of reasoning and tool calling. 

Recently, DeepSeek~\citep{liu2025deepseek}, MiroThinker~\citep{team2025mirothinker} and LongCat~\citep{longcat2026flashthinking} have demonstrated the potential of scaling the depth of agent-environment interactions, \ie, increasing steps of thinking and tool calling to improve deep research capabilities. 
Consequently, recent works have focused on scaling context length or implementing smarter context management strategies to handle deeper reasoning and more tool calls.
On the other hand, while many proprietary models (\eg, GPT-5, Gemini, Claude) support parallel tool calling, no work has yet explored the potential of scaling along the width dimension, \ie, making multiple tool calls in a single step, for deep research agents.

Existing works have explored alternative ways to scale the workload within each step. For example, Kimi-K2.5~\citep{moonshot2026kimik25} introduced a multi-agent framework to deploy multiple sub-agents to solve sub-tasks in a parallel manner; LongCat~\citep{longcat2026flashthinking} introduced parallel reasoning to generate multiple reasoning paths at the same turn and aggregate the outcomes via summarization. However, these approaches either rely on complex orchestration or overlook the coordination of distinct tool executions. In contrast, since parallel tool calling invoke multiple tools within a single reasoning step, it facilitates effective collaboration among the tool calls, improving the efficiency of information gathering.

In this work, we introduce the Wide and Deep (W\&D) research agent to investigate the behavior and performance of state-of-the-art LLMs when jointly scaling depth (via more iterations) and width (via parallel tool calling). 
Moreover, we explore different tool call scheduling strategies to further boost agent performance on deep research tasks. In summary, we make the following contributions: 
\begin{itemize} 
    \item We demonstrate that scaling width via parallel tool calling not only increases performance, but also reduces the number of turns required for the agent to find the correct answer. 
    \item We analyze the driving factors behind why parallel tool calling improves performance by presenting several case studies. 
    \item We study multiple tool call schedulers by employing different numbers of tool calls in each turn to further boost performance, suggesting the potential of dynamic tool calling. 
\end{itemize}

\section{Methodology}

\subsection{Parallel Tool Calling}

\begin{figure}[t]
    \centering
    \includegraphics[width=0.9\linewidth]{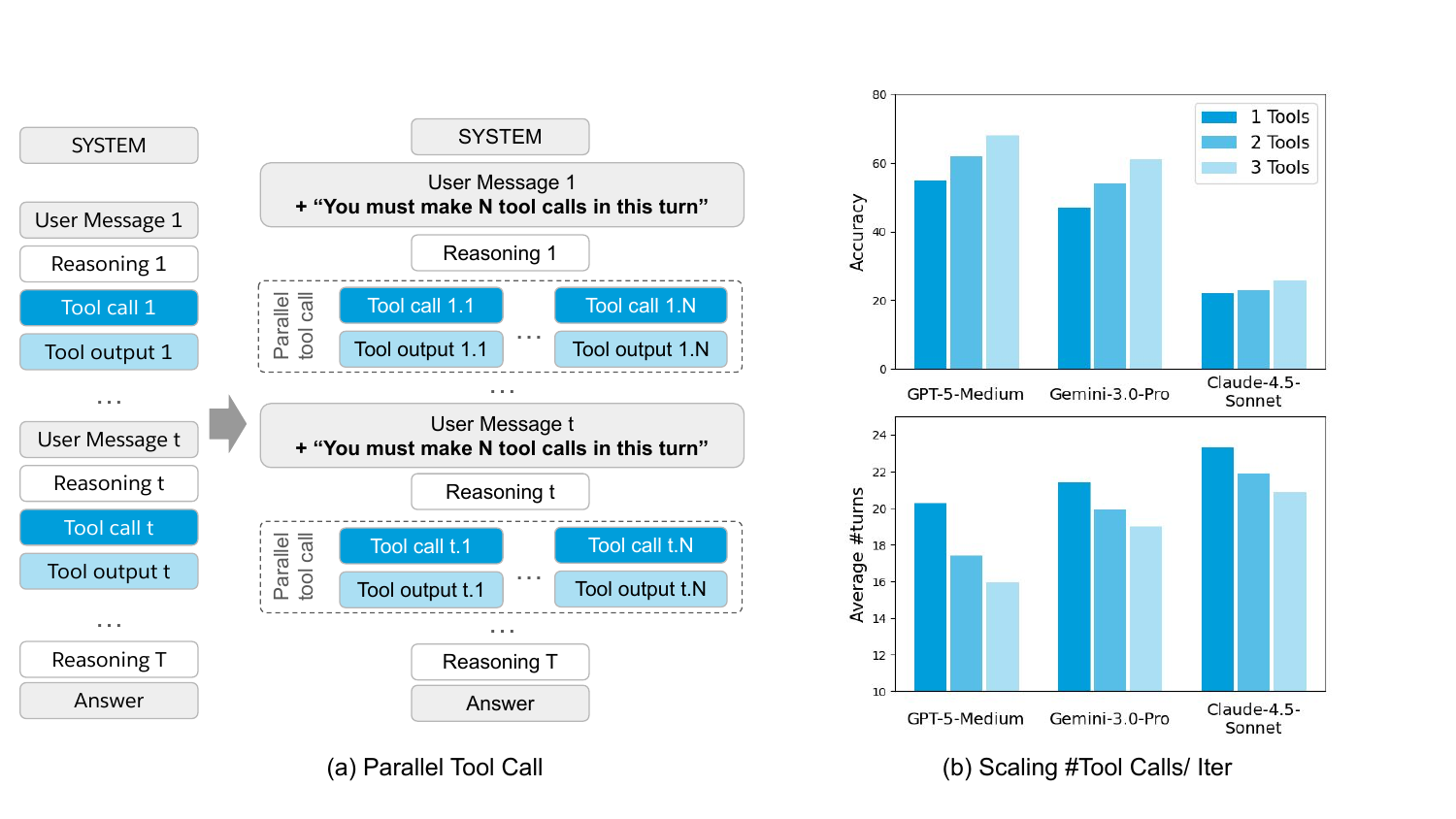}
    \caption{(a) Single vs. parallel tool calling in a multi-step deep research agent trace. In parallel tool calling, the model performs a single reasoning step to issue multiple tool calls simultaneously; these calls are executed in parallel and their outputs are returned together into the agent's trace. (b) Top: Performance of different LLMs under parallel tool calling with varying \# tool calls per step. Performance consistently improves as the \# parallel tool calls increases across all models. Bottom: Average \# turns required to complete the task with different \# parallel tool calls. Increasing the \# tool calls per iteration reduces the total \# iterations needed to complete the deep research task.}
    \label{fig:main_figure}
\end{figure}

State-of-the-art LLMs increasingly support the capability of \textit{parallel tool calling}~\footnote{\url{https://platform.openai.com/docs/guides/function-calling?api-mode=chat\#parallel-function-calling}}~\footnote{\url{https://ai.google.dev/gemini-api/docs/function-calling?example=meeting\#parallel_function_calling}}~\footnote{\url{https://platform.claude.com/docs/en/agents-and-tools/tool-use/implement-tool-use\#parallel-tool-use}}. Denote the LLM as $f_{\theta}$. To define this formally, let us first consider the standard sequential formulation of the agent trace. Given a user query $X$ (which contains system prompt and the user question), at each step\footnote{Throughout this work, the terms `step,' `iteration,' and `turn' are used interchangeably.} $t$ of the agent trace, a typical LLM outputs a reasoning thought $R_t$ and a single tool call $A_t$. The environment executes this call and returns an observation $O_t$. The agent repeats this process until the final step $T$, where it produces a final answer $\hat{Y}$ instead of a tool call. Consequently, the full sequential agent trace $\tau_{\text{seq}}$ is defined as an ordered sequence:
\begin{equation}
\begin{aligned}
    \tau_{\text{seq}} &= \left\langle X, (R_1, A_1, O_1), \dots, (R_{T-1}, A_{T-1}, O_{T-1}), (R_T, \hat{Y}) \right\rangle \\
    R_t, A_t &= f_{\theta}(\left\langle X, (R_1, A_1, O_1), \dots, (R_{t-1}, A_{t-1}, O_{t-1})\right\rangle)
\end{aligned}
\end{equation}

Parallel tool calling extends this paradigm by allowing the agent to generate multiple tool calls simultaneously. At step $t$, rather than issuing a single action $A_t$, the model generates a set of $m$ concurrent tool calls $\mathcal{A}^{\text{par}}_t = \{A^{(1)}_t, \dots, A^{(m)}_t\}$. These are executed in parallel by the environment, yielding a corresponding set of observations $\mathcal{O}^{\text{par}}_t = \{O^{(1)}_t, \dots, O^{(m)}_t\}$. The parallel agent trace $\tau_{\text{par}}$ is thus formalized as:
\begin{equation}
    \tau_{\text{par}} = \left\langle X, (R_1, \mathcal{A}^{\text{par}}_1, \mathcal{O}^{\text{par}}_1), \dots, (R_{T-1}, \mathcal{A}^{\text{par}}_{T-1}, \mathcal{O}^{\text{par}}_{T-1}), (R_T, \hat{Y}) \right\rangle
\end{equation}

\Cref{fig:main_figure} shows the difference in agent trace between the single tool calling and parallel tool calling. This approach allows the agent to aggregate significantly more information per interaction turn, thereby reducing the total number of steps required to solve the task. Furthermore, parallel execution offers dual efficiency benefits compared to performing $m$ sequential steps. First, it amortizes the computational cost of reasoning; we condense $m$ distinct reasoning traces into a single $R_t$, significantly reducing the number of decoding tokens and LLM generation latency. Second, because tool calls within the set $\mathcal{A}_t$ are executed concurrently, the wall-clock time spent waiting for environment feedback is minimized. In summary, parallel tool calling optimizes both the end-to-end latency of the agent rollout and the cost of LLM API usage by reducing both the iteration count and total token consumption.

\textbf{Precise control of tool calling.} To ensure the agent calls the exact number of tools specified, we explored two prompting strategies: 1) adding an instruction in the system message (query $X$) requesting $m$ function calls per iteration, and 2) inserting a user message before each LLM call that specifies the required number of function calls. We chose the latter approach for its superior performance and more reliable tool-call consistency. Specifically, the agent trace is now represented as:

\begin{equation}\tau_{\text{par}} = \left\langle X, U_1, (R_1, \mathcal{A}^{\text{par}}_1, \mathcal{O}^{\text{par}}_1), \dots, U_{T-1}, (R_{T-1}, \mathcal{A}^{\text{par}}_{T-1}, \mathcal{O}^{\text{par}}_{T-1}), U_T, (R_T, \hat{Y}) \right\rangle\end{equation}

The detailed instruction used in $U_t$ is illustrated in \Cref{fig:prompt_parallel_tool}.

\begin{figure}[h]
\begin{tcolorbox}[colback=gray!5, colframe=gray!50, title=User Instruction at Each Iteration]
\small
\itshape
At next step, if you need to make function calls, you MUST make at least \textbf{m} but not more than \textbf{m+1} function calls in a single response to gather information extensively.
\end{tcolorbox}
\caption{Prompt for controlling the number of tool calls in parallel tool calling.}
\label{fig:prompt_parallel_tool}
\end{figure}

\section{Experimental results}

\subsection{Settings}
\textbf{Benchmarks.} We conduct our experiments across three widely adopted deep-research benchmarks: 1) BrowseComp \citep{wei2025browsecomp}, 2) Humanity's Last Exam (HLE) \citep{phan2025humanity}, and 3) GAIA \citep{mialon2023gaia}. Because the evaluated models are not multimodal, we restrict our testing to the text-only subsets of HLE (2,158 samples) and GAIA (103 samples). 
Furthermore, due to the high computational costs of full evaluations, performance on BrowseComp and HLE is reported based on the first 100 samples unless otherwise specified. We evaluate GPT-5 (Medium reasoning effort), Gemini 3.0 Pro, and Claude 4.5 Sonnet across these benchmarks.

\textbf{Tool Environment.} We adopt the agent framework from MCP-Universe~\citep{mcpuniverse} for evaluation. We provide the agent with three tools: 1) a Google-based search tool\footnote{We use Serper API for search}, 2) a scraping tool\footnote{We use JINA API for scraping} with LLM summarization, and 3) a Python code interpreter. The scraping tool accepts a target URL and an extraction query as input; it scrapes the full content of the webpage and passes it, along with the query, to a summarization LLM to extract the specific information required. We use Gemini-2.5-Flash (with thinking mode disabled) for the summary model due to its balance of affordability and effectiveness. For BrowseComp and GAIA, we expose only the search and scraping tools since these benchmarks focus exclusively on information seeking, whereas for HLE, we expose all three tools to the agent. More implementation details can be found in the Appendix.

\subsection{Main result}

\begin{figure}[ht!]
    \centering
    \includegraphics[width=0.67\linewidth]{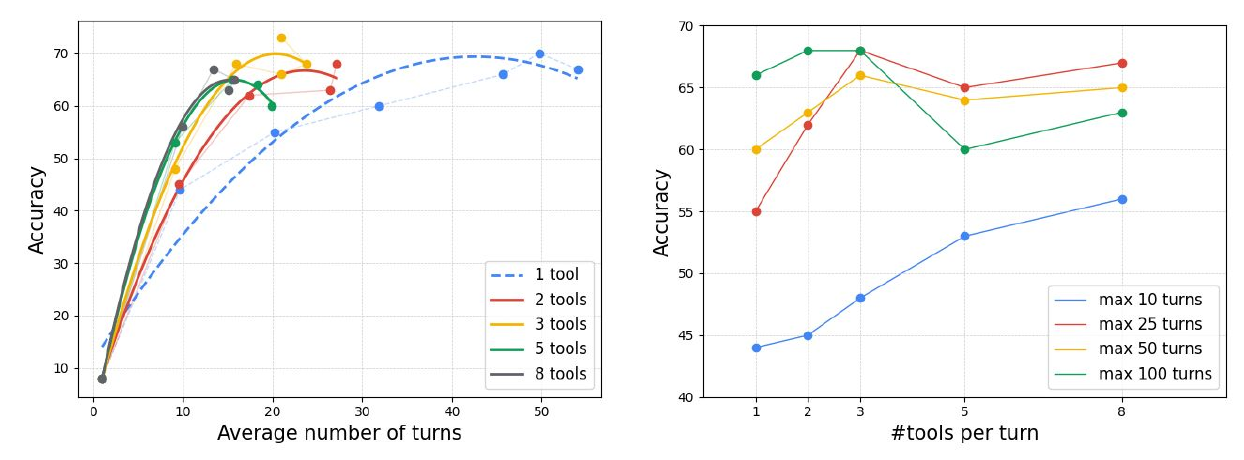}
    \caption{(Left) BrowseComp accuracy against average number of turns. (Right) Accuracy against number of tools per turn. }
    \label{fig:browse_comp_performance_vs_turns_tools}
\end{figure}

\textbf{Tool call scaling.} \Cref{fig:browse_comp_performance_vs_turns_tools} shows the performance of the deep research agent on BrowseComp. Specifically, \Cref{fig:browse_comp_performance_vs_turns_tools} (a) shows the performance of scaling depth (i.e., average \# turns via max turn limit) for both single tool calling and parallel tool calling. \Cref{tab:main_result_table} presents the tabular version of \Cref{fig:browse_comp_performance_vs_turns_tools} (a). The results indicate that parallel tool calling achieves the best performance with $3$ tools per turn with better performance than the single tool calling. Meanwhile, the average number of turns for parallel tool calling shifts to the left, suggesting that parallel tool calling significantly reduces the number of turns required to finish the task. This significantly reduces the wall-clock time to finish the task and the LLM API costs due to the lower number of turns. For example, in single tool calling, to achieve $66$\% accuracy, it costs $\$102.5$ for $100$ tasks in tool calling and LLM API fees and takes an average of $1522.6$ s to finish the trace; whereas for parallel tool calling with 3 tool calls per turn, the model achieves $68$\% accuracy with a cost of $\$65.7$ for $100$ tasks (\ie~, a $35.9\%$ reduction) and a wall-clock time of $904.2$ s (\ie~, a $40.6\%$ reduction). 

\Cref{fig:browse_comp_performance_vs_turns_tools} (b) shows the performance scaling of width (\ie~, \# tool calls per turn via parallel tool calling) under different max turn limits. The results suggest that with a lower max turn limit (\eg~, 10), scaling the width consistently improves performance, whereas with a larger limit (\eg~, 100), the best performance is achieved with a moderate number of tool calls. This suggests that for larger max turn limits, a higher number of tool calls does not always help. Consequently, varying tool calls across different turns might further improve performance. Inspired by this, we provide a further study in~\Cref{sec:tool_call_scheduler}.

\begin{figure}[ht!]
    \centering
    \includegraphics[width=\linewidth]{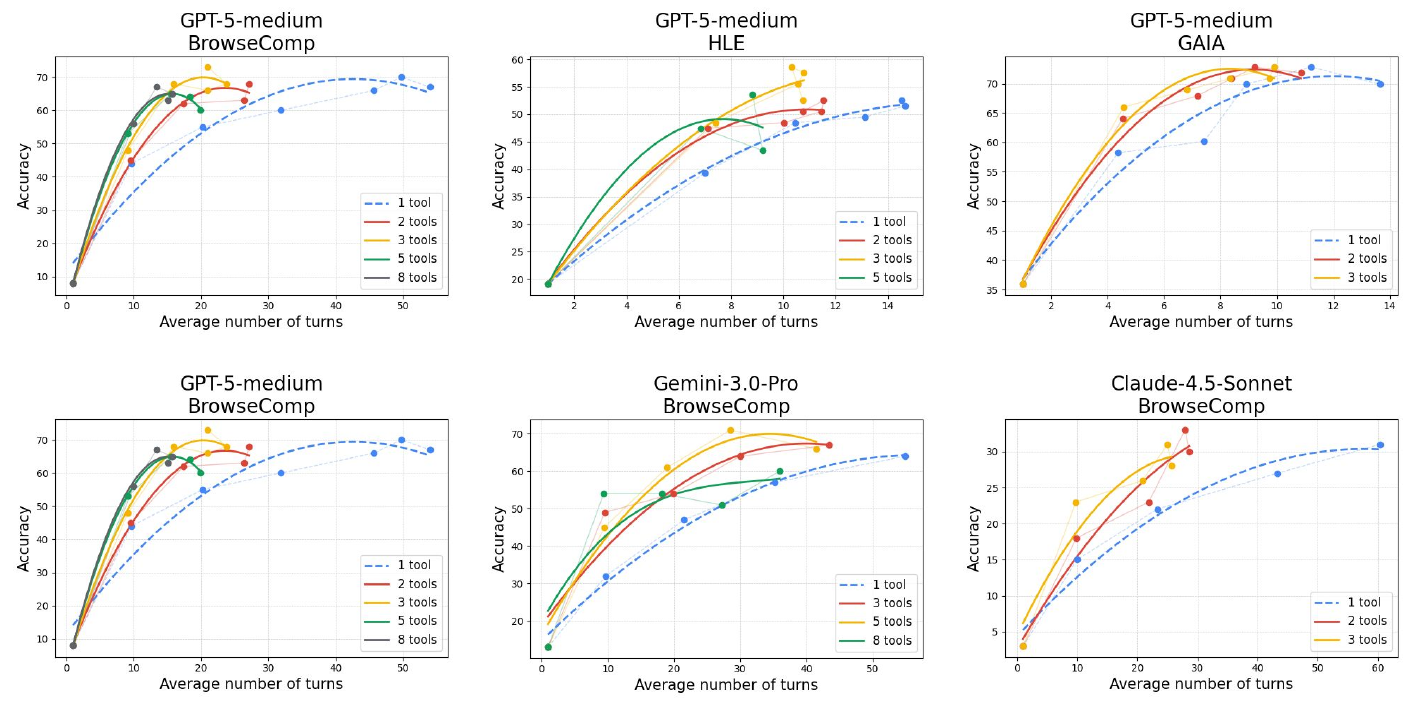}
    \caption{\textbf{Scaling of tool calls.} (Top row) Performance of GPT-5-medium across different benchmarks. (Bottom row) Performance of different models on BrowseComp benchmark.}
    \label{fig:tool_call_generalization}
\end{figure}

\textbf{Generalization on different models and benchmarks.} \Cref{fig:tool_call_generalization} shows the performance of parallel tool calling across different datasets (top row) and different models. The results suggest that the superior performance of parallel tool calling and the benefit of reducing the \# turns required to finish the task are generalizable to different LLMs and deep research benchmarks. To further validate whether the performance gain can be obtained for open-source models, we test the performance of DeepSeek-V3.2~\citep{liu2025deepseek} and Qwen-3-235B-A22B-Thinking-2507~\citep{qwen3technicalreport}. \Cref{tab:open_source_model} shows that there is a small gain in performance when using parallel tool calling; however, the gain is not as prominent as in SoTA proprietary models, suggesting room for improvement in open-source model training to enable more effective parallel tool calling.

\textbf{Full set evaluation.} Our results above for BrowseComp are evaluated with the first 100 tasks. We run the experiment with parallel tool calling on the full set of BrowseComp and obtained 62.2\% accuracy with GPT-5-Medium which is 7.3\% performance gain compared to the GPT-5-High (54.9\%\citep{openai2025gpt5}).

\begin{table}[ht!]
    \centering
    \caption{Accuracy and average number of iterations to completion (in brackets) on the BrowseComp dataset across different tool call limits per iteration. \textbf{No tool call} implies the LLM answers solely via reasoning without tools. \textbf{$n$ iters} indicates the agent is forced to answer and stop at the $n$-th iteration.}\label{tab:main_result_table}
    \begin{tabular}{lccccccc}
        \toprule
        & \textbf{No tool call} & \textbf{10 iters} & \textbf{25 iters} & \textbf{50 iters} & \textbf{100 iters} & \textbf{150 iters} & \textbf{300 iters} \\ 
        \midrule
        \textbf{1 Tool} & & 44 \blue{(9.7)} & 55 \blue{(20.3)} & 60 \blue{(31.9)}  & 66 \blue{(45.7)}  & 70 \blue{(49.8)}  & 67 \blue{(54.7)} \\
        \textbf{2 Tools} &       & 45 \blue{(9.6)} & 62 \blue{(17.4)} & 63 \blue{(26.5)} & \textbf{68} \blue{(27.1)} & 66 \blue{(26.2)}  & - \\
        \textbf{3 Tools} & 8 \blue{(1.0)} & 48 \blue{(9.2)} & \textbf{68} \blue{(16.0)} & \textbf{66} \blue{(21.0)} & \textbf{68} \blue{(23.8)} & 73 \blue{(21.0)} & - \\
        \textbf{5 Tools} &       & 53 \blue{(9.1)} & 65 \blue{(15.6)} & 64 \blue{(18.4)}  & 60 \blue{(19.9)} &    -        & - \\
        \textbf{8 Tools} &       & \textbf{56} \blue{(10.0)} & 67 \blue{(13.4)} & 65 \blue{(15.7)} & 63 \blue{(15.2)} &    -       & - \\
        \bottomrule
    \end{tabular}
\end{table}

\begin{table}[ht!]
    \centering
    \caption{Accuracy and average number of iterations to completion (in brackets) on the BrowseComp dataset for open-source models with single tool calling and parallel tool calling.}\label{tab:open_source_model}
    \begin{tabular}{lccccccc}
        \toprule
        & Single tool calling & Parallel tool calling (3 tools) \\ 
        \midrule
        Qwen3-235B-A22B-Thinking-2507 & 8 \blue{(18.5)} & 11 \blue{(52.1)} \\
        DeepSeek-V3.2 & 38 \blue{(78.0)} & 39 \blue{(52.5)} \\
        \bottomrule
    \end{tabular}
\end{table}
\section{Why Parallel Tool Calling Improves Accuracy}
Our previous experiments have shown both the efficiency and the effectiveness of parallel tool calling. While the efficiency improvement is easier to understand due to the decreased number of iterations and reduced reasoning, the source of the effectiveness (i.e., the gain in performance) remains unclear. To understand better on why the parallel tool calling helps effectiveness we inspect the agent trace manually to gain more insight and identified the following 3 patterns.

\paragraph{Observation 1: Exploration improves the credibility of information sources.} In information-seeking tasks, the credibility of the source is vital for accuracy. Parallel tool calling broadens the search scope by triggering multiple queries, thereby aggregating a diverse collection of sources. This allows the model to compare inputs during the reasoning phase and select the most authoritative one. For the question ``According to United Nations data from 2021, what percentage of national parliamentary seats in Northern Africa were held by women?'', the parallel tool calling retrieved statistics from multiple locations and explicitly selected an official UN report, leading to the correct answer. In contrast, the single-tool calling, limited by a narrower search scope, relied on a API database. This source proved unofficial, causing the agent to answer incorrectly. The detailed comparison is shown in~\Cref{fig:information_source}.

\begin{figure}[ht!]
\begin{tcolorbox}[colback=white, colframe=black, sharp corners, boxrule=1pt]
    \textit{Question: According to United Nations data from 2021, what percentage of national parliamentary seats in Northern Africa were held by women?}\newline
    
    \textbf{Single tool calling}\newline    
    What I did: ... Queried the SDG API ...\newline
    Evidence and result: The API returns a record with geoAreaName: Northern Africa ... value: 22.25 ...\newline
    Conclusion: According to United Nations SDG API data for 2021, women held 22.25\% of national parliamentary seats in Northern Africa ...\newline
    
    \textbf{Parallel tool calling}\newline
    I searched United Nations sources and located the UN Statistical Yearbook tables ... Both the SYB65 (2022) and SYB67 (2024) tables list Northern Africa’s percentage by year and show 24.3\% for 2021 ... The UN Statistical Yearbook explicitly gives 24.3\% for Northern Africa in 2021.
\end{tcolorbox}
\caption{The comparison between single tool calling vs parallel tool calling in using the information source. Parallel tool calling uses more reliable information source.}
\label{fig:information_source}
\end{figure}

\begin{figure}[ht!]
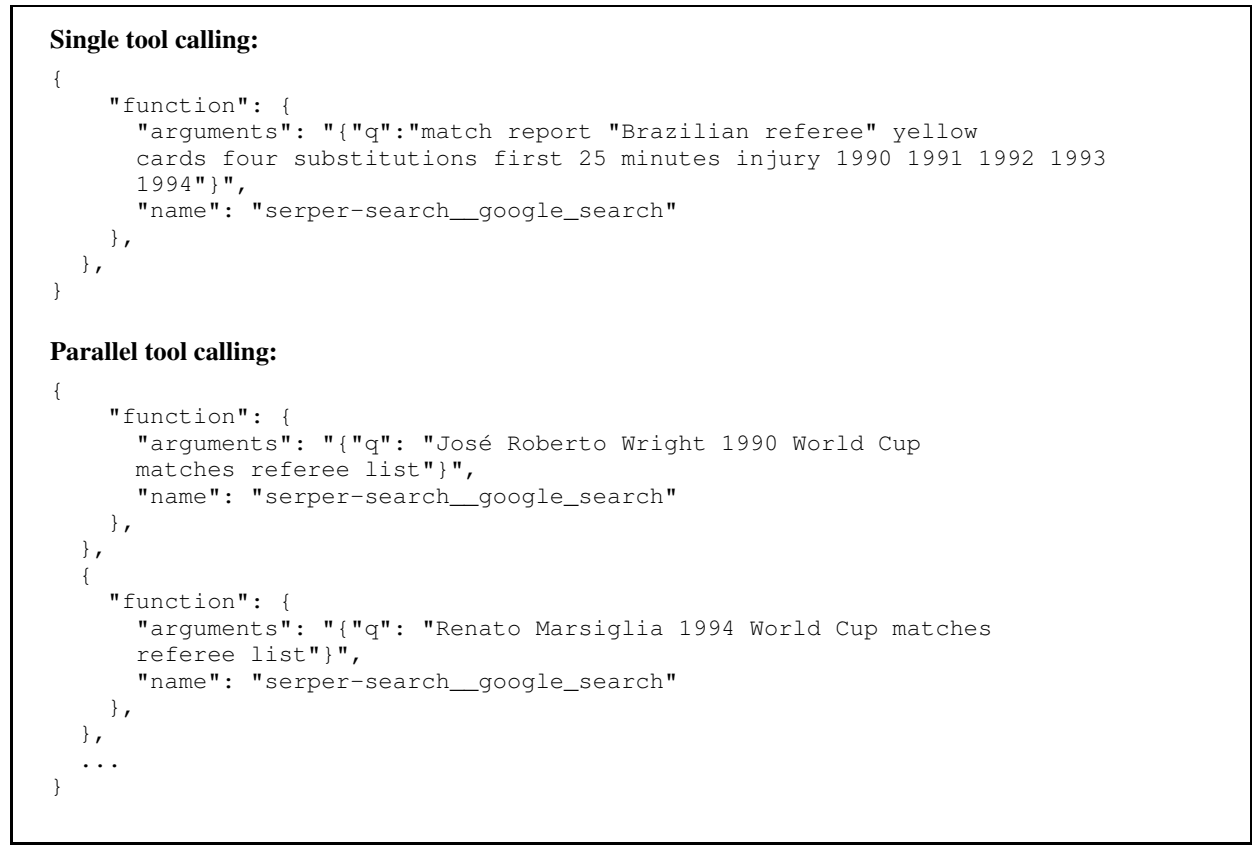

\begin{tcolorbox}[colback=white, colframe=black, sharp corners, boxrule=1pt]
    \textbf{Single tool calling:}
    \begin{small}
    \begin{verbatim}
{
    "function": {
      "arguments": "{"q":"match report "Brazilian referee" yellow 
      cards four substitutions first 25 minutes injury 1990 1991 1992 1993 
      1994"}",
      "name": "serper-search__google_search"
    },
  },
}
    \end{verbatim}
    \end{small}

\textbf{Parallel tool calling:}
\begin{small}
\begin{verbatim}
{
    "function": {
      "arguments": "{"q": "José Roberto Wright 1990 World Cup 
      matches referee list"}",
      "name": "serper-search__google_search"
    },
  },
  {
    "function": {
      "arguments": "{"q": "Renato Marsiglia 1994 World Cup matches 
      referee list"}",
      "name": "serper-search__google_search"
    },
  },
  ...
}
    \end{verbatim}
    \end{small}
\end{tcolorbox}
\caption{The comparison between single tool calling vs parallel tool calling in using the search tool. Parallel tool calling decompose a complex query to multiple simpler ones which enables more effective search.}
\label{fig:search_tool}
\end{figure}

\paragraph{Observation 2: Tool call redundancy enables tool results verification and avoids unreliable tools.} When making tool calls, we could encounter tool call failure or unreliable tool call output. For agent with single tool call it will take all the tool call results as it is and plan the subsequent steps. Therefore, when there is a unreliable tool call result, the agent will be mis-led by this result and hence head toward the wrong direction of problem solving. In parallel tool calling, the agent tends to make redundant tool calls by specifying different arguments for each tool to get similar information. As a result, these different tool call results serve as verification of the results with each other and hence the model can decide whether this tool results are reliable or not and decide to use this information or do further searching. This verification ensure that the agent will not use the unrelible tool results and increased performance. In a query regarding Virginia Tech tuition, both agents located a target PDF, but the scraping mechanism failed to retrieve content. The tool’s internal summarization model, however, hallucinated an answer based on this empty input. The single-call agent blindly accepted this fabricated tool output as fact. In contrast, the parallel-call agent triggered multiple extraction attempts. Because the resulting hallucinations were inconsistent across the redundant calls, the agent identified the results as unreliable, initiated a new search, and successfully retrieved the correct data.
\paragraph{Observation 3. Parallel search enhances retrieval effectiveness through query decomposition.} While single-tool agents often struggle to return relevant results for complex, multi-faceted constraints, parallel tool calling allows the model to decompose a complicated request into multiple, simpler search queries. This approach significantly improves the search engine's ability to recall specific information. Consider a complex query regarding soccer matches between 1990–1994 involving specific constraints on the referee, yellow cards, and injury substitutions. The single tool calling attempted a keyword-stuffed search: ``match report Brazilian referee yellow cards four substitutions first 25 minutes injury 1990... 1994''. This yielded poor results. In contrast, the parallel tool calling split the task into distinct, manageable queries, such as "1990 World Cup matches referee list" and "1994 World Cup matches referee list". By isolating the search variables, the parallel approach achieved higher recall and successfully retrieved the necessary data. The detailed comparison is shown in~\Cref{fig:search_tool}.

\section{Tool call scheduler}\label{sec:tool_call_scheduler}

In the previous section, we fixed the number of tool calls across all steps. However, this may not be the optimal strategy. As shown in the right plot of \Cref{fig:browse_comp_performance_vs_turns_tools}. We can see that when number of steps is small, increasing number of tools per step improves accuracy; when number of steps is higher, having more tools may not be beneficial. In this section, we present a preliminary exploration of different tool call schedulers for parallel tool calling. Denoting $m_t$ as the number of tool calls required at step $t$, we compare the following schedulers:

\begin{itemize}
    \item \textbf{Constant 1 Tool}: The default setting where make single tool call at each step ($m_t=1$)
    \item \textbf{Constant 3 Tools}: We fix the number of tool calling to 3 across all steps ($m_t=3$)
    \item \textbf{Ascending}: The number of tool calls increases \textit{w.r.t.} step. \begin{equation}
        m_t = \begin{cases} 
        1 & t \leq 25\\
        2 & 25 < t \leq 50\\
        3 & t > 50
        \end{cases}
    \end{equation}
    \item \textbf{Descending}: The number of tool calls decreases. \textit{w.r.t.} step. \begin{equation}
        m_t = \begin{cases} 
        3 & t \leq 25\\
        2 & 25 < t \leq 50\\
        1 & t > 50
        \end{cases}
    \end{equation}
    \item \textbf{Automatic}: We let LLM to decide the number of tool calls at each step. We add a specific instruction to the user message at each step, as shown in \Cref{fig:prompt_dynamic_parallel_tool}.
\end{itemize}

\begin{figure}[h]
    \centering
    \begin{tcolorbox}[colback=gray!5, colframe=gray!50, title=User Instruction at Each Iteration for Dynamic Tool Calling]
        \small
        \itshape
        "In the next step, first identify your progress (0-100\%) on solving this problem. If you need to make function calls, you MUST make at least 1 but no more than 4 function calls in a single response to gather information extensively. Based on the progress, the general principle is to make more function calls in the early phases, and gradually reduce the number as you approach completion."
    \end{tcolorbox}
    \caption{The specific instruction used to let the LLM decide the number of tool calls autonomously.}
    \label{fig:prompt_dynamic_parallel_tool}
\end{figure}

\begin{figure}[ht!]
    \centering
    \includegraphics[width=0.9\linewidth]{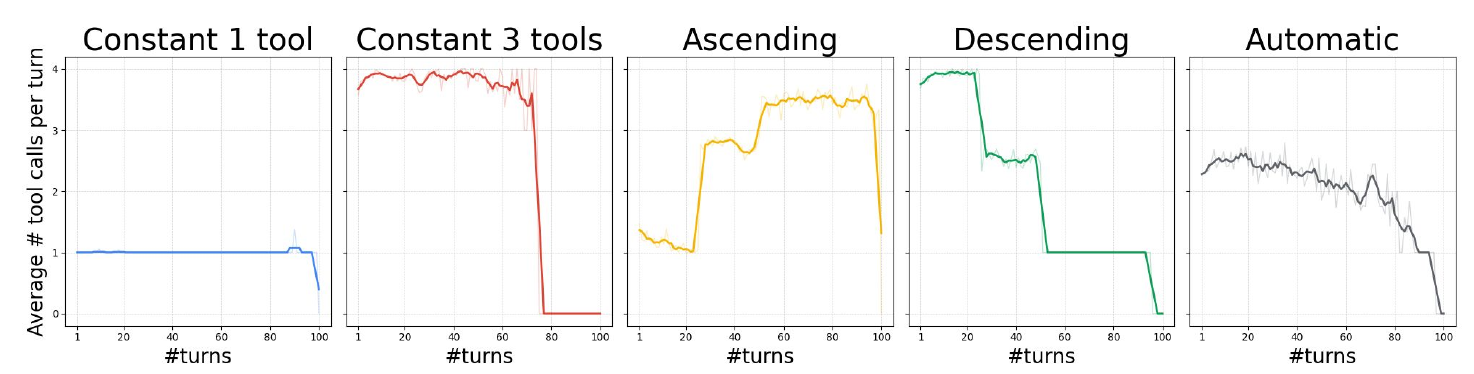}
    \caption{The average \# tool calls across all turns for different tool call scheduler.}
    \label{fig:tool_scheduler_call_stats}
\end{figure}

\Cref{fig:different_schedulers} shows the performance across different tool call schedulers. The results show that \textbf{Descending} achieves a $6\%$ performance gain over \textbf{Constant 3 Tools}, demonstrating the potential of tool call schedulers. \textbf{Ascending} achieves the worst performance while \textbf{Descending} achieves the best, suggesting that the explore-then-exploit strategy contributes to better performance. The \textbf{Automatic} strategy did not perform better than \textbf{Descending}, indicating that the LLM itself cannot determine the optimal number of tool calls in each iteration. These results could inspire future research on training LLMs to incorporate the explore-then-exploit tool call scheduler, enabling them to decide the optimal number of tool calls independently. \Cref{fig:tool_scheduler_call_stats} shows the actual average number of tool calls in each turn. The number of tool calls is very close to our specified tool calls, suggesting that the LLMs follow our instruction in~\Cref{fig:prompt_parallel_tool} well.

\begin{table}[ht!]
    \centering
    \caption{Accuracy and average number of iterations to completion (in brackets) on the BrowseComp dataset across different tool call schedulers.}\label{fig:different_schedulers}
    \begin{tabular}{lccccc}
        \toprule
        & \textbf{Constant 1 Tool} & \textbf{Constant 3 Tools} & \textbf{Ascending} & \textbf{Descending} & \textbf{Automatic}  \\ 
        \midrule
        Accuracy \blue{(\#Turns)} & 66 \blue{(45.7)} & 68 \blue{(23.8)} & 63 \blue{(36.5)} & \textbf{74} \blue{(23.5)} & 72 \blue{(26.6)} \\
        \bottomrule
    \end{tabular}
\end{table}

\section{Related work}
Existing works have explored similar ideas to handle a larger workload within each step. They can be categorized into two main classes: 1) parallel reasoning, which generates multiple reasoning traces based on the same context and employs a summary model to aggregate all the results; and 2) sub-agent design, where the main agent invokes multiple sub-agents in each step to complete sub-tasks and aggregates the results at the end of the step.

\textbf{Parallel reasoning.} The work of~\citep{pan2025learning} proposes to replace serialized chain-of-thought reasoning with coordinated parallel reasoning to improve latency; however, it does not target the agent setting where tool calls are incurred. Longcat~\citep{longcat2026flashthinking} introduced parallel reasoning to generate multiple reasoning paths within the same turn, aggregating the outcomes via summarization. However, it overlooks coordination between different paths and hence can produce repetitive work. In contrast, in our parallel tool calling, although all tools are invoked in the same step, they share the same reasoning, which helps to coordinate between tool calls to avoid the wasteful use of tools.

\textbf{Sub-agent.} Miroflow~\citep{2025miroflow} proposes a sub-agent orchestration design. In each iteration, the main agent reasons based on the current context without direct tool calling; instead, a sub-agent is invoked to use tools and solve sub-tasks. Kimi-K2.5~\citep{moonshot2026kimik25} proposes an agent swarm framework and trains the model with the newly proposed parallel-agent reinforcement learning (PARL). The model learns to decompose the task into parallelizable sub-tasks. These sub-agent-based methods are similar to parallel tool calling; however, they require more complex orchestration and significant modification to the commonly used single-agent framework. In contrast, our parallel tool calling enables more workload to be processed and is readily integrated into any single-agent framework, making it potentially adaptable to any agentic system.

\section{Conclusion}

In this work, we introduced the \textbf{Wide and Deep research agent}, a framework designed to explore the benefits of scaling execution width via parallel tool calling alongside reasoning depth. Our extensive experiments across BrowseComp, HLE, and GAIA demonstrate that parallel tool calling significantly improves agent performance while simultaneously reducing the number of sequential turns required to reach a solution. This reduction in iterations not only lowers the end-to-end wall-clock time but also amortizes reasoning costs by condensing multiple actions into single reasoning steps. Through qualitative analysis, we identified that these gains are driven by enhanced source verification, redundancy against tool failures, and effective query decomposition. Furthermore, our investigation into tool call scheduling reveals that a \textbf{Descending} strategy, prioritizing broad exploration in early stages followed by focused exploitation, outperforms static or ascending strategies. However, the inability of current LLMs to autonomously optimize this trade-off in the \textbf{Automatic} setting highlights a limitation in existing models. Future work should focus on training agents, potentially through reinforcement learning, to dynamically manage the ``width-depth'' trade-off, enabling next-generation agents to autonomously navigate the explore-then-exploit spectrum for high-efficiency deep research.

\section*{Acknowledgments}

We thank the Salesforce AI Research team for their valuable feedback and support throughout this project. 
\bibliography{references}
\bibliographystyle{ieeetr}

\appendix
\section{Appendix}\label{appendix}

\textbf{More implementation details.} \Cref{fig:system_prompt} provides the full system prompt we use in the deep research agent. In each step, as mentioned in~\Cref{fig:prompt_parallel_tool}, a user message is used to precisely control the number of tool calls required in the next step. Furthermore, we include a countdown message to inform the LLM of the remaining budget (see~\Cref{fig:user_count_down}). This is inspired by previous work~\citep{anthropic2025claudeopus45} that uses prompts to make the LLM aware of the remaining budget and hence improves performance. When the agent reaches the max turn limit, we use a user message (see~\Cref{fig:user_force_result}) to force the LLM to output the answer given all the information available, instead of stopping without an answer.

\begin{figure}[h!]
\begin{tcolorbox}[colback=gray!5, colframe=gray!50, title=User Instruction at Each Iteration]
\small
\itshape

You are an intelligent assistant that can solve complex problems by thinking step-by-step and using available tools when needed.

Today is:

\{\{FORMATTED\_DATE\}\}

\textbf{Your Role}

\{\{INSTRUCTION\}\}

\textbf{Your Task}

\{\{QUESTION\}\}

\textbf{How You Work}
\begin{enumerate}
    \item \textbf{Think First}: Analyze the problem and determine what information or actions you need
    \item \textbf{Use Tools When Needed}: Call appropriate functions to gather information, perform calculations, or take actions
    \item \textbf{Reason with Results}: Process the tool outputs and use them to inform your next steps
    \item \textbf{Iterate}: Continue thinking and using tools until you can provide a complete answer
\end{enumerate}

\textbf{Available Capabilities}
\begin{itemize}
    \item You have access to various specialized tools through function calling
    \item When you need to use a tool, simply call the appropriate function with the required parameters
    \item The system will execute the function and provide you with the results
    \item Use these results to continue your problem-solving process
\end{itemize}

\textbf{Important Guidelines}
\begin{itemize}
    \item You have a maximum of \{\{MAX\_STEPS\}\} steps to complete this task
    \item Each step should either advance your understanding or gather necessary information
    \item Be systematic and thorough in your approach
    \item Only provide your final answer when you have sufficient information
    \item \textbf{IMPORTANT: At each step, you MUST follow user instructions in each step to make the amount of function calls specified in the user instructions.}
\end{itemize}

---

\textbf{Final Answer Format}
When you have completed your analysis and gathered all necessary information, provide your final response using this JSON format:

\begin{verbatim}
{
    "thought": "Explain your reasoning process and how you arrived at the answer",
    "answer": "Your complete final answer to the task"
}
\end{verbatim}

\textbf{Important}: 
\begin{itemize}
    \item Use the JSON format above ONLY for your final answer
    \item During your thinking process, you can respond in any natural format
    \item The "answer" field should contain your complete solution as a string
\end{itemize}
\end{tcolorbox}
\caption{System prompt for deep research agent.}
\label{fig:system_prompt}
\end{figure}

\begin{figure}[ht!]
\begin{tcolorbox}[colback=gray!5, colframe=gray!50, title=User Instruction at Each Iteration]
\small
\itshape
You have $n$ steps remaining. Please continue. At next step, if you need to make function calls, you MUST make at least 3 but not more than 4 function calls in a single response to gather information extensively.
\end{tcolorbox}
\caption{User countdown message.}
\label{fig:user_count_down}
\end{figure}

\begin{figure}[ht!]
\begin{tcolorbox}[colback=gray!5, colframe=gray!50, title=User Instruction at Each Iteration]
\small
\itshape
Important: You have reached the maximum number of interaction turns. Therefore, you have to output the final answer with the specified format without calling any tools in the final response.
\end{tcolorbox}
\caption{User message to force the LLM to generate the final answer when it reaches the max turn limit.}
\label{fig:user_force_result}
\end{figure}

\end{document}